\documentclass[lettersize,journal]{IEEEtran}

\usepackage{amsmath,amsfonts,amssymb}
\usepackage{graphicx}
\usepackage{cite}
\usepackage{array}
\usepackage{booktabs}
\usepackage{multirow}
\usepackage{makecell}
\usepackage{threeparttable}
\usepackage{url}
\usepackage{xcolor}
\usepackage[hidelinks]{hyperref}

\begin{document}

\title{\LARGE \bf AE-UAV: An Air-to-Air Event Camera Dataset for UAV Detection and Tracking with Continuous-Time Annotations}

\author{Zixin Jiang,
    Bing He,
    Chaoran Xiong, \emph{Graduate Student Member, IEEE},
    Zhenzhen Wang,
    \\ Xin Zhao\(^*\),
    and Ling Pei\(^*\), \emph{Senior Member, IEEE}
    \thanks{This work was supported in part by the National Natural Science Foundation of China (NSFC) under Grant No. 62273229, and in part by the Science and Technology Commission of Shanghai Municipality under Grant Nos. 24DZ3101300, 24TS1402600, and 24TS1402800. (Corresponding author: Ling Pei)}
    \thanks{Zixin Jiang, Bing He, Zhenzhen Wang and Xin Zhao are with Department of Electronic Engineering, Rocket Force University of Engineering, Xi’an 710025, China (e-mail: jiangzixin0214@163.com; hebing202301@163.com; wangzhenzhen202404@163.com; zhaoxin20062111@163.com).}
    \thanks{Chaoran Xiong and Ling Pei are with Shanghai Key Laboratory of Navigation and Location Based Services, Shanghai Jiao Tong University, Shanghai 200240, China (e-mail: sjtu4742986@sjtu.edu.cn; ling.pei@sjtu.edu.cn).}
    \thanks{Ling Pei is also with State Key Laboratory of Submarine Geoscience, School of Automation and Intelligent Sensing, Shanghai Jiao Tong University, Shanghai 200240, China.}
}

\maketitle
\thispagestyle{empty}
\pagestyle{empty}

\begin{abstract}
Perceiving unmanned aerial vehicles (UAVs) from another UAV is essential for aerial avoidance, interception, and counter-UAV operations. Event cameras suit this task because of their microsecond temporal resolution and high dynamic range. However, existing event-based UAV datasets were mostly recorded by static or ground-based observers, and their targets are annotated only at fixed frame rates. We present AE-UAV, an air-to-air event camera dataset collected from a flying UAV for detection and tracking. It contains 178 sequences of event data at 1280×720 resolution covering daytime and nighttime flights under three motion geometries. A continuous-time annotation framework based on cubic B-spline interpolation provides ground-truth boxes at arbitrary timestamps and a label for every event. Hold-out experiments confirm a median interpolation error of 2.19 pixels, reducing the piecewise-constant baseline error by 74\%. We benchmark six detectors and six trackers across illumination and motion conditions. The results show that nighttime and head-on scenarios are the hardest cases for detectors. A cross-rate generalization study further shows that learned trackers lose their accuracy when the test aggregation rate differs from the training rate. The dataset will be publicly released.
\end{abstract}

\begin{IEEEkeywords}
Data Sets for Robotic Vision, Aerial Systems: Perception and Autonomy, Visual Tracking, Object Detection, Segmentation and Categorization, Event Camera
\end{IEEEkeywords}

\section{Introduction}
\label{sec:introduction}

As unmanned aerial vehicles (UAVs) increasingly share low-altitude airspace, perceiving other UAVs from a flying platform has become essential for avoidance, interception, and counter-UAV operations~\cite{anti_uav,detfly}. Air-to-air (A2A) observation poses challenges distinct from those of ground-based surveillance, including dual-platform relative motion, small distant targets, and dense background clutter induced by observer ego-motion. Conventional frame cameras provide neither sufficient frame rates nor adequate dynamic range for the fast motion and extreme illumination changes~\cite{event_camera_survey}.

\begin{figure}[!t]
\centering
\includegraphics[width=\columnwidth]{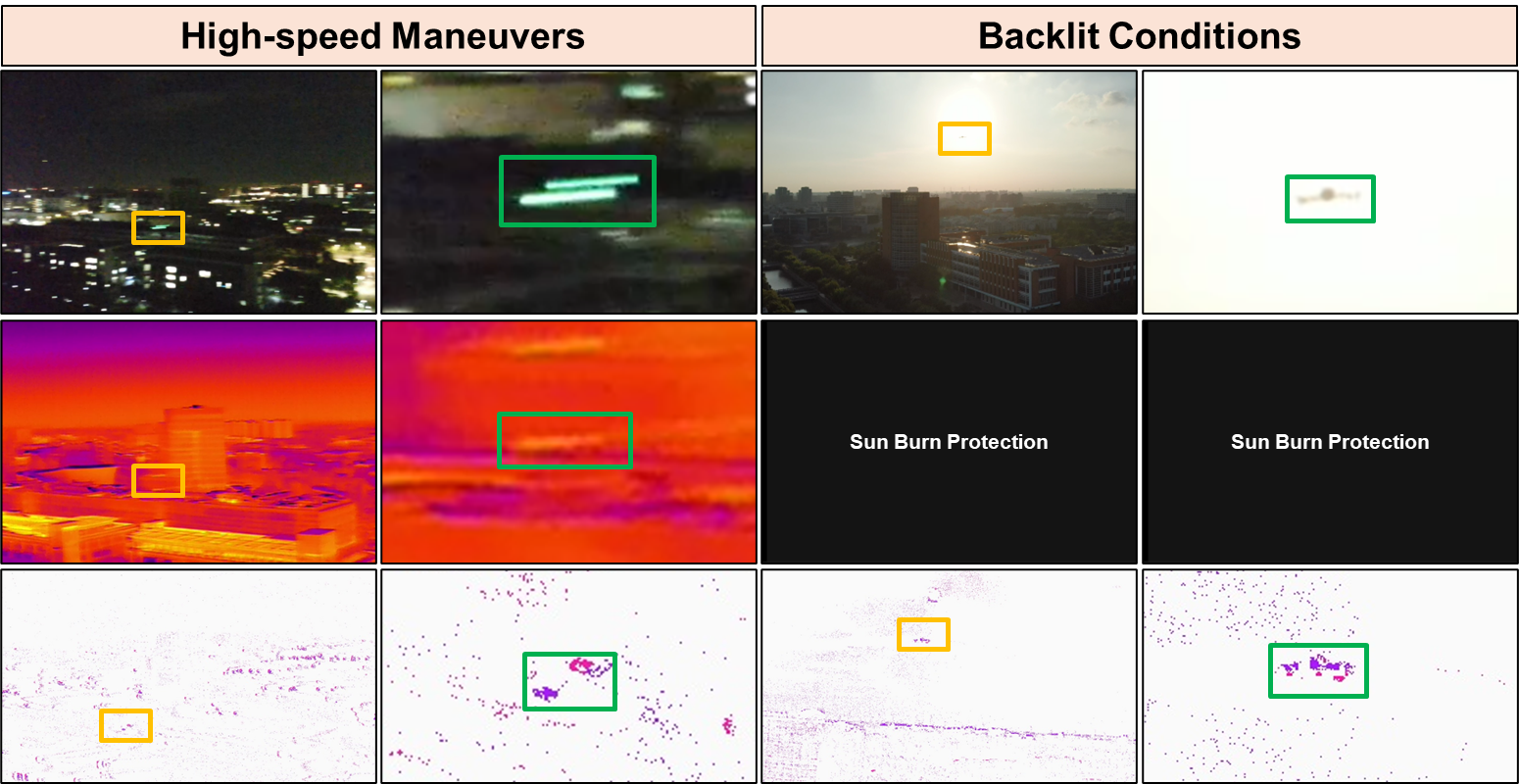}
\caption{Qualitative comparison of RGB, thermal, and event imaging under challenging A2A scenarios. Columns 2 and 4 are magnified views of the yellow regions in columns 1 and 3, respectively.}
\label{fig:modality_comparison}
\end{figure}

Event cameras report per-pixel brightness changes asynchronously with microsecond resolution and high dynamic range~\cite{low_latency_event}. These advantages have been exploited on mobile robots for quadrotor avoidance~\cite{falanga2020} and perception on cars, drones, and legged robots~\cite{mvsec,dsec,mseviq}. 
To illustrate these advantages in A2A conditions, Fig.~\ref{fig:modality_comparison} presents a qualitative comparison across three imaging modalities. 
Under high-speed maneuvers, RGB cameras produce severe motion blur. Under backlit conditions, the thermal camera's sunburn-protection mechanism blanks out the image.
In contrast, event cameras maintain consistent target visibility in both cases. These characteristics make event cameras well suited for A2A perception, though realizing this potential requires representative benchmarks.

Despite growing interest in event-based UAV observation, two key limitations persist across existing datasets. First, most datasets rely on static or ground-based platforms~\cite{fuavd,nerdd,evuav,evflying,fred,mmuav,skyev}, where background events are sparse and motion alone often reveals the target. The few datasets~\cite{evdetmav, m2euav} that employ a moving observer cover only daytime conditions, leaving nighttime scenarios unaddressed despite the operational importance of night surveillance. 
Second, existing labels are produced at a fixed frame rate. Event-level labels hold each box constant over its frame window~\cite{evuav}. When a method is evaluated at a different temporal resolution, the fixed-rate labels no longer describe the target at the evaluated instants.

We present AE-UAV, an event camera dataset for A2A UAV detection and tracking collected from a flying UAV. It covers daytime and nighttime flights, and it provides continuous-time ground truth that can be queried at any timestamp. The main contributions are as follows:
\begin{itemize}
    \item An A2A event camera dataset. AE-UAV contains 178 sequences of $1280\times720$ event data. It covers three motion geometries and three illumination conditions.
    \item A continuous-time annotation framework based on cubic B-spline interpolation. It generates ground truth at arbitrary timestamps and labels for every event. A hold-out experiment quantifies the interpolation accuracy.
    \item Detection and single-object tracking benchmarks. We evaluate representative detectors and trackers across multiple scenarios, and use the continuous-time ground truth to evaluate trackers at aggregation rates that differ from the training rate, revealing that accuracy drops substantially as the test rate increases.
\end{itemize}

\section{Related Work}
\label{sec:related}

\subsection{Event Camera Datasets on Mobile Robots}
\label{sec:rw-robot}
The high temporal resolution and dynamic range of event cameras suit platforms under rapid ego-motion. MVSEC~\cite{mvsec} first deployed stereo event cameras on UAVs, cars, and motorcycles with lidar and motion-capture ground truth. DSEC~\cite{dsec} raised the resolution and synchronization quality for driving, and M3ED~\cite{m3ed} extended multi-sensor event capture to cars, quadrotors, and legged robots. CEAR~\cite{cear} and EvTTC~\cite{evttc} further addressed agile quadruped locomotion and vehicular time-to-collision estimation, respectively. These datasets supply ground truth for ego-state, scene depth, or collision time. AE-UAV complements them with ground truth for another aerial robot observed from a flying robot.

\subsection{Datasets for UAV Detection and Tracking}
\label{sec:rw-uav}
Frame-based datasets have established UAV detection and tracking as a recognition problem. Det-Fly~\cite{detfly} evaluated deep detectors on small UAVs photographed from another UAV, and ARD-MAV~\cite{ardmav} added cluttered backgrounds with abrupt camera motion. Anti-UAV410~\cite{anti_uav} contributed a large-scale thermal infrared tracking benchmark from a ground observer. All three rely on frame cameras with fixed rates and limited dynamic range.

General-purpose event tracking benchmarks such as VisEvent~\cite{visevent} and EventVOT~\cite{eventvot} include UAVs only as a minor category. Dedicated datasets followed. F-UAV-D~\cite{fuavd} and NeRDD~\cite{nerdd} record static-camera event streams for UAV detection. FRED~\cite{fred} and EV-Flying~\cite{evflying} extend the setting to RGB-event fusion and multiple flying object categories, respectively. EV-UAV~\cite{evuav} introduces per-event piecewise-constant labels for tiny target segmentation. MM-UAV~\cite{mmuav} combines RGB, thermal, and event modalities for multi-UAV tracking, and SkyEV~\cite{skyev} adds camera motion through a ground rig. All these datasets use fixed or ground-level observers and annotate bounding boxes on cumulative event frames at a constant rate.

A few recent efforts move the event camera onto a mobile platform. EventMAV~\cite{evdetmav} introduces the moving-camera dataset, covering three MAV types, and exploits propeller periodicity for detection. ODD-SEC~\cite{oddsec} spins an event camera on a ground vehicle for omnidirectional detection. The closest prior work is M$^2$E-UAV~\cite{m2euav}, which records airborne event and inertial data for daytime detection. Following EV-UAV~\cite{evuav}, its per-event labels are derived from boxes annotated at 10 Hz.
AE-UAV advances this setting in two ways. It adds nighttime flights, including sequences with target navigation lights off, and covers pursuit, evasion, and head-on geometries. It supplies continuous-time ground truth via cubic B-spline interpolation and quantifies the accuracy of this temporal model. Table~\ref{tab:dataset_comparison} summarizes the comparison.

\begin{table*}[!t]
\centering
\vspace*{5pt}
\caption{Comparison of Event Camera Datasets for UAV Detection and Tracking}
\label{tab:dataset_comparison}
\resizebox{\textwidth}{!}{%
\begin{tabular}{l l c c c c l c}
\toprule
\textbf{Dataset} & \textbf{Observer} & \textbf{Event res.} & \textbf{\makecell[c]{UAV\\Specific}} & \textbf{\makecell[c]{Night/\\Low-light}} & \textbf{\makecell[c]{Annotation\\Method}} & \textbf{Tasks} & \textbf{Year} \\
\midrule
VisEvent~\cite{visevent}  & Ground   & $346\times260$  & \(\times\)     & \(\checkmark\) & Cumulative Frame  & SOT & 2024 \\
EventVOT~\cite{eventvot}  & Ground   & $1280\times720$ & \(\times\)     & \(\checkmark\) & Cumulative Frame                          & SOT & 2024 \\
F-UAV-D~\cite{fuavd}      & Static   & $1280\times720$ & \(\checkmark\) & \(\times\) & Cumulative Frame                         & Det & 2024 \\
NeRDD~\cite{nerdd}        & Static   & $1280\times720$ & \(\checkmark\) & \(\times\) & Cumulative Frame                         & Det & 2024 \\
EV-UAV~\cite{evuav}       & Static   & $346\times260$  & \(\checkmark\) & \(\checkmark\) & Events (piecewise constant)    & Det/Seg & 2025 \\
EV-Flying~\cite{evflying} & Static   & $1280\times720$ & \(\times\)     & \(\times\) & Cumulative Frame                           & Det & 2025 \\
FRED~\cite{fred}          & Static   & $1280\times720$ & \(\checkmark\) & \(\checkmark\) & Cumulative Frame                         & Det/MOT & 2025 \\
MM-UAV~\cite{mmuav}       & Ground   & $346\times260$  & \(\checkmark\) & \(\checkmark\) & Cumulative Frame        & MOT/SOT & 2025 \\
EventMAV~\cite{evdetmav}  & Moving   & $640\times480$  & \(\checkmark\) & \(\times\) & Cumulative Frame               & Det & 2025 \\
SkyEV~\cite{skyev}        & Ground/Moving & $1280\times720$ & \(\checkmark\) & \(\times\) & Cumulative Frame       & Det/MOT & 2026 \\
ESOT500~\cite{stare}      & Ground   & $346\times260$  & \(\times\) & \(\times\) & 500 Hz Cumulative Frame              & SOT & 2026 \\
M\(^2\)E-UAV~\cite{m2euav}  & Airborne & $1280\times720$ & \(\checkmark\) & \(\times\) & Events (propagated) & Det & 2026 \\
\midrule
\textbf{AE-UAV (ours)}    & \textbf{Airborne} & $1280\times720$ & \(\checkmark\) & \(\checkmark\) & \makecell{\textbf{Cumulative Frame +}\\\textbf{Events (cubic)}} & Det/SOT & 2026 \\
\bottomrule
\end{tabular}}
\par\vspace{2pt}{\begin{minipage}{\textwidth}
\footnotesize
Observer: Static = fixed camera; Ground = handheld or ground rig; Moving = moving camera; Airborne = camera on a flying UAV. Night / low light: the dataset contains nighttime or explicitly labeled low-light sequences. SOT/MOT = single/multiple object tracking; Det = detection; Seg = event-level segmentation.
\end{minipage}}
\end{table*}

\subsection{Event-Based Detection, Tracking, and Temporal Ground Truth}
\label{sec:rw-methods}

Event-specific detectors convert events into dense tensors and add temporal memory. RVT~\cite{rvt} uses recurrent vision transformers, S5-ViT~\cite{s5vit} replaces recurrence with state space models, GET~\cite{get} groups events into tokens, and SMamba~\cite{smamba} exploits sparsity to reduce computation. For tracking, transformer-based trackers designed for RGB video~\cite{ostrack,romtrack,aqatrack,odtrack} can be applied to rendered event frames, and FocusTrack~\cite{focustrack} adapts local sampling to counter-UAV tracking. Event-specific designs include MambaEVT~\cite{mambaevt} and eMoE-Tracker~\cite{emoe}.

Most benchmarks annotate boxes on frames rendered at a fixed rate. A method that runs at another rate or reports results at other instants must then be scored against labels that were not defined for those instants. Streaming perception~\cite{streaming} and the STARE framework with the 500~Hz ESOT500 dataset~\cite{stare} showed that ignoring latency and annotation rate overstates online accuracy. ESOT500 addresses the problem by dense annotation. AE-UAV instead obtains ground truth at arbitrary instants from sparse keyframes by interpolation and measures the error this step introduces.

\section{Acquisition System and Plan}
\label{sec:platform}

\subsection{Platforms and Recording}
\label{sec:sensor}

\begin{figure}[!t]
\centering
\includegraphics[width=0.9\columnwidth]{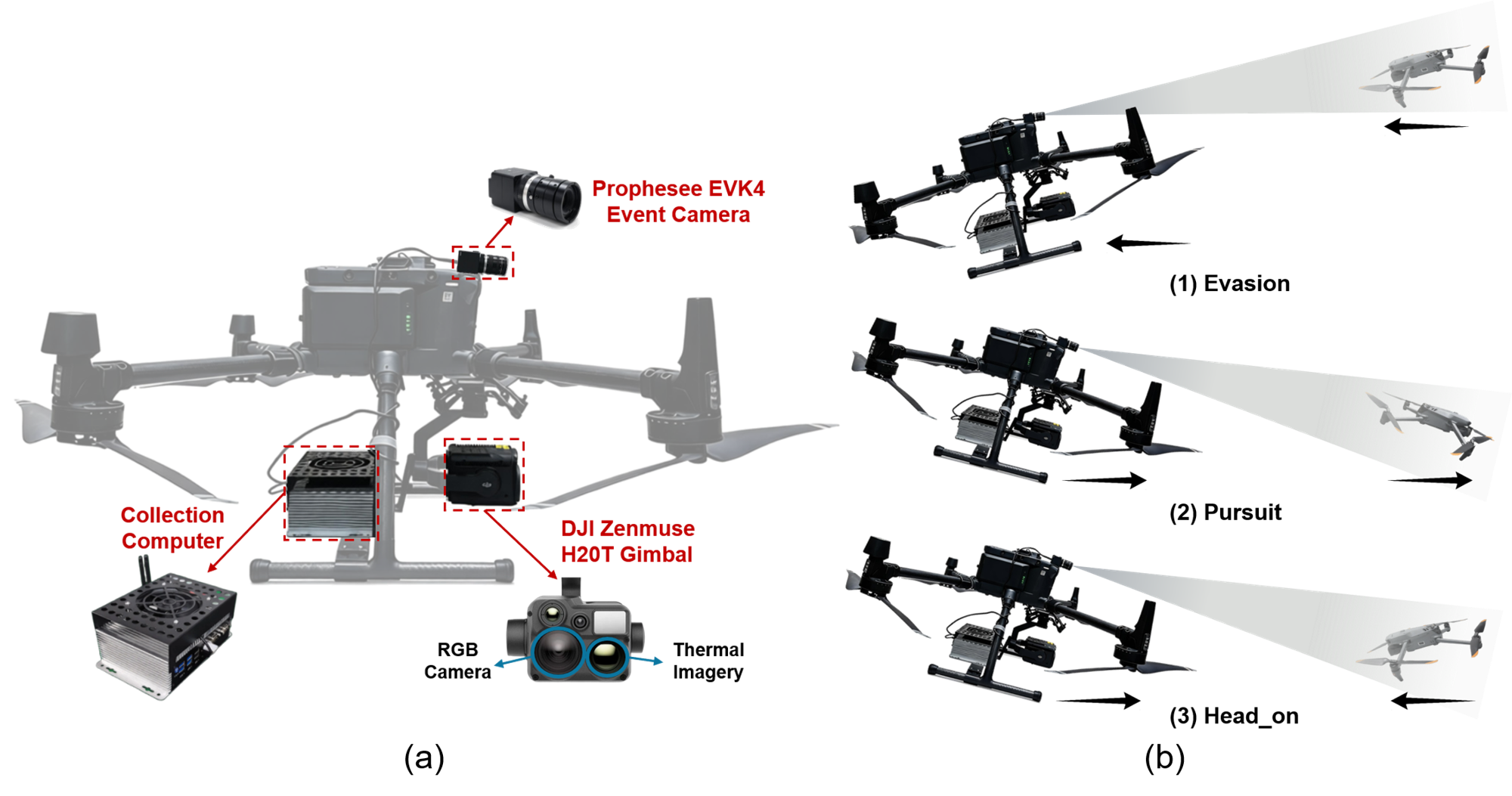}
\caption{Data acquisition. (a) Acquisition system configuration. (b) The three motion geometry patterns: pursuit, evasion, and head-on approach.}
\label{fig:collect}
\end{figure}

The acquisition system employs two coordinated UAVs for systematic capture of realistic aerial scenarios. 
The observer platform is a DJI Matrice~300~RTK that carries a Prophesee EVK4~HD event camera and an onboard collection computer, as shown in Fig.~\ref{fig:collect}(a). The event camera is rigidly mounted on the airframe, so the full ego-motion of the observer appears in the event stream. This camera incorporates a Sony IMX636 sensor with $1280\times720$ resolution and is fitted with a 12~mm fixed-focus lens. A DJI Zenmuse~H20T gimbal additionally records RGB and thermal images, but these images are not synchronized with the events and serve only for illustration. AE-UAV is therefore an event-only dataset, and we focus on this setting in order to explore what can be perceived from events alone. The target is a DJI Mavic~3T, a 920~g enterprise quadrotor with a 380.1~mm diagonal wheelbase, which is representative of typical small UAVs in surveillance and counter-UAV applications.

Events were recorded at full sensor resolution through the Prophesee ROS driver on an Intel Core~i7 computer with a 2~TB solid-state drive.
Sensor biases were kept at factory defaults. Both the event-rate controller and the anti-flicker filter were disabled to preserve the raw sensor output without any hardware-level event suppression.
All timestamps and annotations refer to the microsecond sensor clock, and per-sequence checks of inter-event intervals and event-rate saturation found no data loss. Because AE-UAV contains only a single event sensor, no cross-sensor synchronization is required.

\subsection{Collection Protocol}
\label{sec:protocol}

The acquisition protocol is organized around three factors that govern how the target appears in A2A event data, as summarized in Table~\ref{tab:scenario}. Each factor is divided into several categories.

Motion geometry determines the relative motion between the target and the background. Three relative motion patterns are captured as shown in Fig.~\ref{fig:collect}(b): pursuit, evasion, and head-on approach. In a pursuit, the observer follows the target, and both platforms fly in the same direction. In an evasion, the target follows the observer, which looks upward at the target. In a head-on approach, the two platforms fly toward each other.

Illumination controls how many events the target and the background produce. Daytime ego-motion converts the textured background into dense background events. At night, the scene darkens and sensor noise increases. We further divide nighttime flights by the state of the target navigation lights. Nighttime sequences account for approximately 78\% of the dataset. 

Target trajectory determines how the target moves in the image plane. The target UAV executes four types of maneuvers: straight-line, turning, lateral, and vertical. Such maneuvers change the image-plane velocity, visible profile, and the background behind the target. The inter-UAV distance ranges from 15 to 100~m, producing target scales from point-like signatures to detailed structures. Backgrounds range from textured urban areas to clear sky.

\begin{table}[t]
\centering
\caption{Scenario Distribution in the AE-UAV Dataset}
\label{tab:scenario}
\renewcommand{\arraystretch}{1.1}
\begin{threeparttable}
\begin{tabular}{@{}llc@{}}
\toprule
\textbf{Factor} & \textbf{Category} & \textbf{Sequences} \\
\midrule
\multirow{3}{*}{Motion Geometry} 
    & Pursuit & 78 \\
    & Evasion & 81 \\
    & Head-on Approach & 19 \\
\midrule
\multirow{3}{*}{Illumination}
    & Daytime & 39 \\
    & Night with Target Lights On & 98 \\
    & Night with Target Lights Off & 41 \\
\midrule
\multirow{4}{*}{Trajectory}
    & Straight-line & 27 \\
    & Turning & 89 \\
    & Lateral Maneuver & 42 \\
    & Vertical Maneuver & 20 \\
\bottomrule
\end{tabular}
\begin{tablenotes}
\footnotesize
\item[] Daytime scenes include backlit scenes.
\end{tablenotes}
\end{threeparttable}
\end{table}

\section{Continuous-Time Ground Truth}
\label{sec:gt}

\begin{figure*}[!t]
\centering
\vspace*{5pt}
\includegraphics[width=0.8\textwidth]{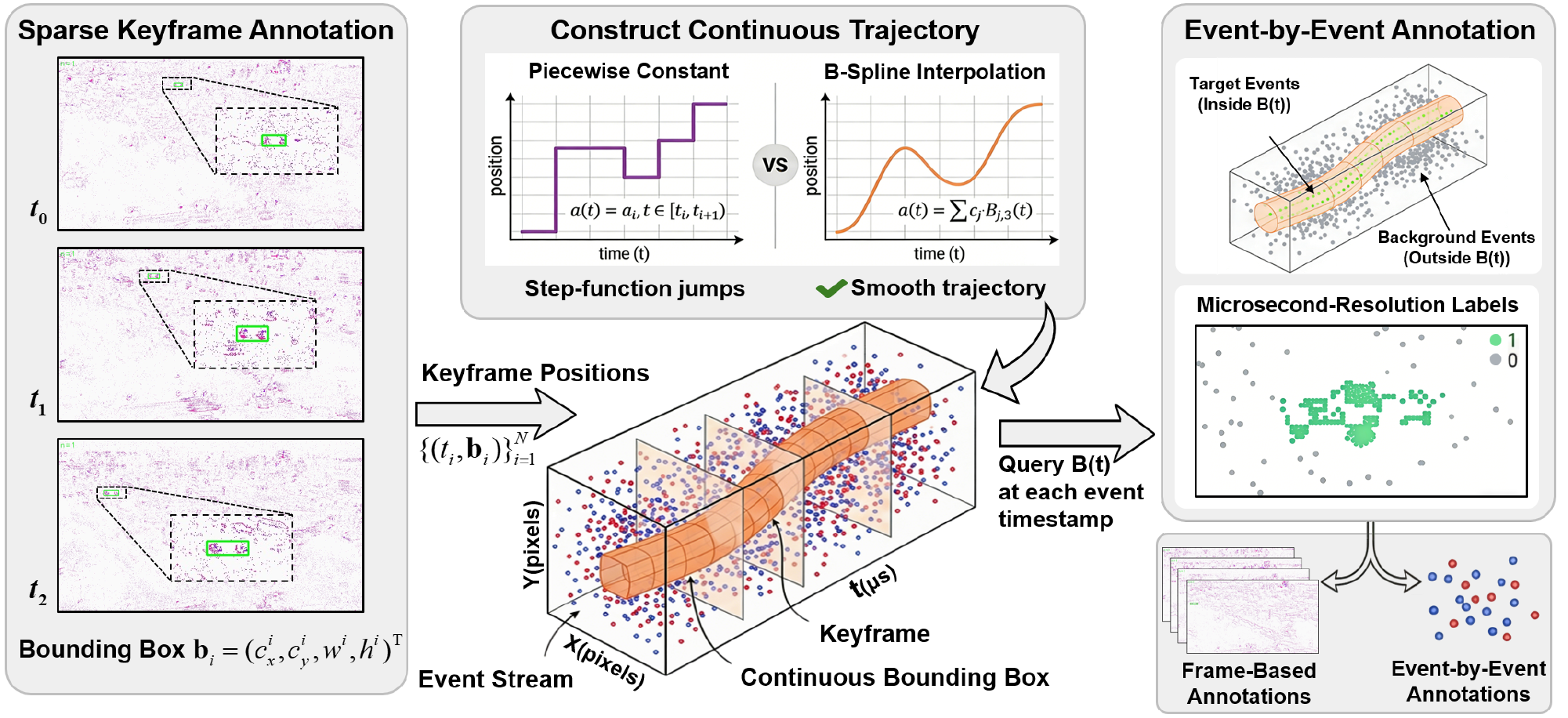}
\caption{Continuous-time annotation pipeline. Left: sparse keyframe bounding boxes. Center: cubic B-spline interpolation constructs $C^2$-continuous trajectories versus piecewise-constant labels. Right: event-by-event labels generated at microsecond resolution.}
\label{fig:annotation}
\end{figure*}
Conventional annotation methods assign constant bounding boxes within inter-frame intervals, creating a mismatch between the continuous nature of event streams and the discrete nature of ground-truth labels. When methods are evaluated at different temporal resolutions, constant bounding-box labels yield inconsistent ground truth, making cross-resolution comparisons unreliable. A continuous annotation framework is therefore essential for event-based benchmarks that aim to support evaluation across varying temporal scales. We propose a continuous-time annotation framework based on cubic B-spline interpolation that bridges sparse keyframe annotations and dense event-level labels. The pipeline operates in three stages as illustrated in Fig.~\ref{fig:annotation}.

\subsection{Keyframe Annotation}
\label{sec:keyframe}
Each sequence is rendered into a video whose $n$-th frame uniformly accumulates all events in the interval $[n/f_k,\,(n+1)/f_k)$, where $f_k = 15$~Hz is the keyframe rate. Annotators label the target in each frame using DarkLabel~2.4, exporting bounding boxes in the format of top-left coordinates, width, and height. Two annotators labeled the data and  reviewed every sequence. Because events within each frame are accumulated uniformly, we assign each keyframe a timestamp at the midpoint of its window,
\begin{equation}
t_i=\frac{n_i+1/2}{f_k},\qquad \mathbf{b}_i=(c_x^i,c_y^i,w^i,h^i)^\top,
\label{eq:keyframe}
\end{equation}
where $n_i$ is the frame index and $(c_x^i,c_y^i)$ is the box center. This yields $N$ keyframes $\{(t_i,\mathbf{b}_i)\}_{i=1}^{N}$ per sequence.

\subsection{Temporal Models}
\label{sec:temporal-model}

A temporal model turns the keyframes into a box $\mathbf{B}(t)$ at any time $t$ covered by the annotated windows, $t\in[t_1-\tfrac{1}{2f_k},\,t_N+\tfrac{1}{2f_k}]$. The simplest model, used for event-level labels in EV-UAV~\cite{evuav}, holds each box over its own window,
\begin{equation}
\mathbf{B}_{\mathrm{PC}}(t)=\mathbf{b}_i,\qquad t\in\left[t_i-\tfrac{1}{2f_k},\;t_i+\tfrac{1}{2f_k}\right).
\label{eq:pc}
\end{equation}
This piecewise-constant (PC) approximation assumes zero intra-frame motion. Such an assumption is frequently violated whenever the target moves between keyframes. In AE-UAV, the apparent speed of the target has a median of 126~px/s. Within one 66.7~ms window, the target therefore moves by about 8~px at the median speed, compared with a median box height of 21~px.

We instead model each box parameter $a\in\{c_x,c_y,w,h\}$ with an interpolating cubic
spline expressed in the B-spline basis,
\begin{equation}
a(t)=\sum_{j=1}^{N}\alpha_j^{a}B_{j,3}(t),\qquad a(t_i)=a_i,\quad i=1,\dots,N,
\label{eq:spline}
\end{equation}
where $B_{j,3}$ are cubic B-spline basis functions and $\alpha_j^{a}$ their coefficients. The curve passes through every keyframe and is twice continuously differentiable, so the box moves smoothly and has no jumps at keyframe boundaries. We keep the human-annotated boxes exact rather than smoothing them, and Section~\ref{sec:gt-accuracy} evaluates a smoothing spline as a reference.

Following the not-a-knot construction, the knot vector repeats $t_1$ and $t_N$ four times and uses $t_3,\dots,t_{N-2}$ as interior knots, so the $N$ keyframes match exactly $N$ basis functions. The coefficients are thus the unique solution of the banded system $\mathbf{A}\boldsymbol{\alpha}^{a}=\mathbf{a}$, with $A_{ij}=B_{j,3}(t_i)$ and $\mathbf{a}=(a_1,\dots,a_N)^\top$.
Because keyframe times and knots are interleaved, the system is non-singular~\cite{deboor}, and the resulting curve passes through every keyframe.
We solve one such system per box parameter per sequence, which requires at least four keyframes per sequence. Within the first and last half-windows, i.e., up to 33~ms before $t_1$ and after $t_N$, the trajectory is extrapolated by the first and last polynomial segments.

\subsection{Event-Level Labels}
\label{sec:event-labels}

Given the fitted trajectory $\mathbf{B}(t) = (c_x(t), c_y(t), w(t), h(t))^\top$, each event $e_m = (x_m, y_m, t_m, p_m)$, is assigned a binary label:
\begin{equation}
\ell_m = 
\begin{cases}
1, & \text{if } (x_m, y_m) \in \mathcal{R}\bigl(\mathbf{B}(t_m)\bigr) \\
0, & \text{otherwise}
\end{cases},
\label{eq:event_label}
\end{equation}
where $\mathcal{R}\bigl(\mathbf{B}(t)\bigr) = \{(x,y) \mid |x - c_x(t)| \leq w(t)/2 \;\wedge\; |y - c_y(t)| \leq h(t)/2\}$ is the axis-aligned rectangular region defined by $\mathbf{B}(t)$.
The pipeline produces two output formats, namely frame-level annotations for conventional trackers and event-level labels for methods operating at native temporal resolution.

\subsection{Annotation Accuracy}
\label{sec:gt-accuracy}

We validate the temporal models through a hold-out experiment. For each sequence, only every $s$-th keyframe is retained with $s \in \{2, 3, 4\}$, corresponding to intervals of 133, 200, and 267~ms. Each model predicts the bounding boxes at the withheld keyframes, and the predictions are compared with the manual annotations. Because the smallest interval is already twice the original keyframe spacing, the measured errors are conservative estimates of the interpolation error at 15~Hz.

Table~\ref{tab:gt_accuracy} presents the results. At 133~ms, the cubic B-spline achieves a median center error of 2.19~px, compared with 8.54~px for the PC model. 
Its mean IoU with withheld keyframes is 0.764, compared with 0.492 for the PC model.
Linear interpolation performs similarly to the cubic B-spline at 133~ms but degrades faster as keyframes become sparser.
From 133 to 267~ms, its median center error grows from 2.31 to 3.47~px, whereas that of the cubic B-spline grows only from 2.19 to 2.86~px, and its mean IoU falls to 0.667 versus 0.710 for the cubic B-spline. 
The smoothing spline is marginally better than the cubic B-spline at 133~ms, but it over-smooths sparse keyframes. Its mean IoU drops below that of linear interpolation at both 200 and 267~ms and reaches 0.633 at 267~ms. Because smoothing offers a negligible gain at the densest tested interval but degrades markedly as keyframes become sparser, we retain the original keyframe values rather than smoothing them.

Table~\ref{tab:gt_speed} breaks down the hold-out error at 133~ms by target apparent speed. The piecewise-constant error grows from 4.50~px below 100~px/s to 24.67~px above 300~px/s. In contrast, the cubic B-spline error remains between 2.12 and 2.28~px across all speed bins, and its advantage over linear interpolation increases with speed. The residual error of the cubic B-spline is therefore dominated by annotation noise rather than target motion.

\begin{table}[!t]
\centering
\vspace*{5pt}
\caption{Hold-Out Accuracy of the Temporal Models}
\label{tab:gt_accuracy}

\setlength{\tabcolsep}{3pt}
\resizebox{\columnwidth}{!}{%
\begin{tabular}{lcccccc}
\toprule
\multirow{2}{*}{Model} & \multicolumn{2}{c}{\textbf{133~ms}} & \multicolumn{2}{c}{\textbf{200~ms}} & \multicolumn{2}{c}{\textbf{267~ms}} \\
\cmidrule(lr){2-3}\cmidrule(lr){4-5}\cmidrule(lr){6-7}
 & CE$\downarrow$ & IoU$\uparrow$ & CE$\downarrow$ & IoU$\uparrow$ & CE$\downarrow$ & IoU$\uparrow$ \\
\midrule
Piecewise constant & 8.54 & 0.492 & 8.73 & 0.485 & 10.51 & 0.423 \\
Linear                                & 2.31 & 0.760 & 2.83 & \underline{0.718} & \underline{3.47} & \underline{0.667} \\
Smoothing spline              & \textbf{2.15} & \textbf{0.765} & \underline{2.58} & 0.715 & 3.50 & 0.633 \\
\textbf{Cubic} & \underline{2.19} & \underline{0.764} & \textbf{2.43} & \textbf{0.745} & \textbf{2.86} & \textbf{0.710} \\
\bottomrule
\end{tabular}}
\par\vspace{2pt}{\begin{minipage}{\linewidth}
\footnotesize
CE: median center error in pixels; IoU: mean IoU with the withheld human box. Smoothing spline: weight selected by generalized cross-validation.
\end{minipage}}
\end{table}

\begin{table}[!t]
\centering
\caption{Hold-Out Error at 133~ms Interval Across Apparent Speeds}
\label{tab:gt_speed}
\setlength{\tabcolsep}{3pt}
\resizebox{\columnwidth}{!}{%
\begin{tabular}{l c c c c | c c c}
\toprule
\multirow{2}{*}{Speed (px/s)} 
  & \multirow{2}{*}{Share}
  & \multicolumn{3}{c}{Median CE (px)\,$\downarrow$} 
  & \multicolumn{3}{c}{IoU $<$ 0.5 (\%)\,$\downarrow$} \\
\cmidrule(lr){3-5}\cmidrule(lr){6-8}
  & & PC & Lin. & Cubic & PC & Lin. & Cubic \\
\midrule
$<$100   & 37.9\% & 4.50  & 2.26 & \textbf{2.21} 
         & 18.2 & 5.7 & \textbf{5.4} \\
100--200 & 37.5\% & 9.49  & 2.25 & \textbf{2.12} 
         & 55.4 & 4.0 & \textbf{3.4} \\
200--300 & 15.4\% & 15.81 & 2.46 & \textbf{2.28} 
         & 85.9 & 4.1 & \textbf{3.4} \\
$>$300   & 9.2\%  & 24.67 & 2.50 & \textbf{2.27} 
         & 96.4 & 4.2 & \textbf{2.9} \\
\bottomrule
\end{tabular}}
\par\vspace{2pt}{\begin{minipage}{\linewidth}
\footnotesize
Share: fraction of the withheld keyframes in each speed bin. PC: piecewise constant; Lin.: linear.
\end{minipage}}
\end{table}

\section{Dataset Statistics and Benchmark Protocol}
\label{sec:stats-protocol}

\subsection{Statistics}
\label{sec:stats}

\begin{figure*}[!t]
\centering
\vspace*{5pt}
\includegraphics[width=0.8\textwidth]{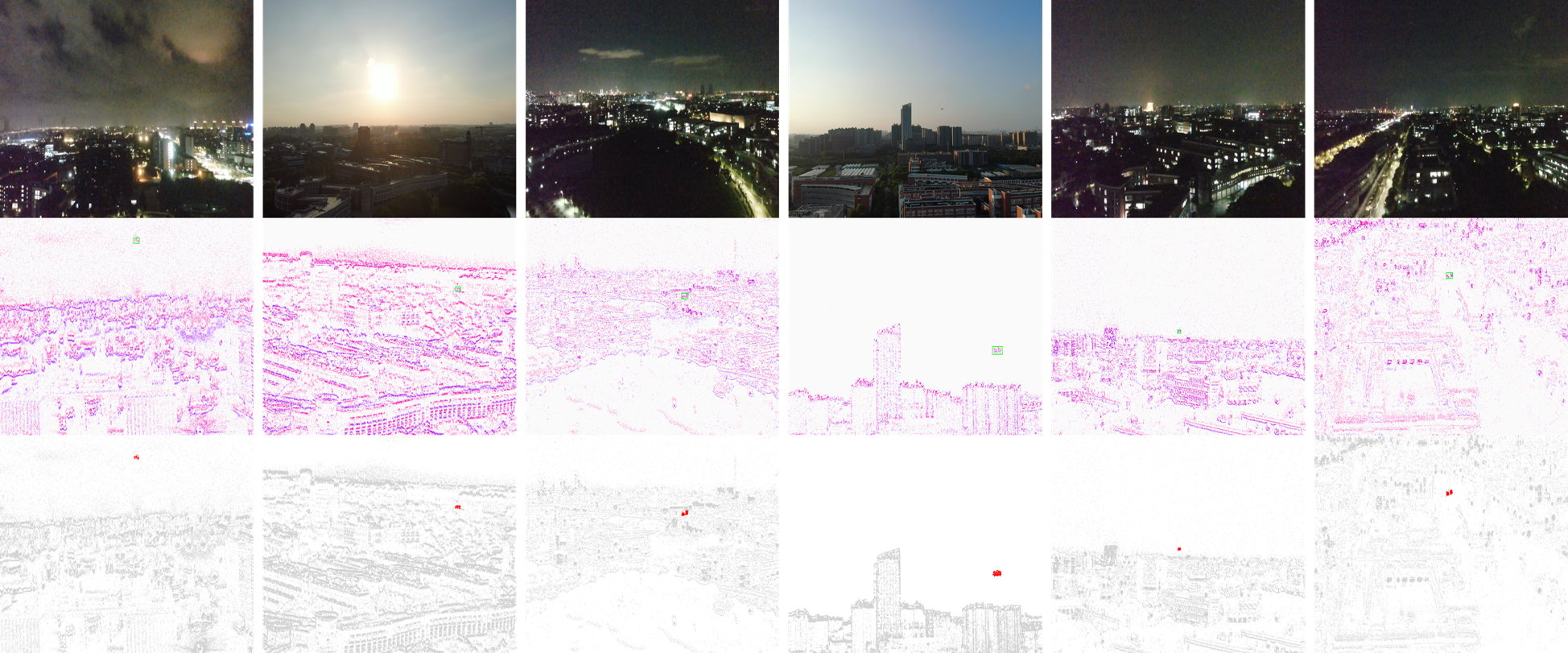}
\caption{Qualitative overview of representative scenes in the AE-UAV dataset. From top to bottom, the rows show the RGB reference image, the annotated event accumulation image, and the labeled raw event points. Each column corresponds to a different scene. The green and red markers indicate the annotated target location.}
\label{fig:dataset_scenes}
\end{figure*}

The AE-UAV dataset comprises 178 sequences totaling approximately 2140~s and over 8.15 billion events, of which 31.2 million are target events accounting for 0.38\% of all events. This extreme class imbalance is inherent in A2A scenarios where the target subtends a small portion of the field of view. Fig.~\ref{fig:dataset_scenes} provides a qualitative overview of the dataset across representative scenes. Each column corresponds to a different scene, the scenes cover urban skylines, nighttime flights, and cluttered backgrounds. This visualization highlights the dense ego-motion events that fill the field of view and the small spatial footprint of the target relative to the surrounding clutter.

\begin{figure}[!t]
\centering
\includegraphics[width=0.85\columnwidth]{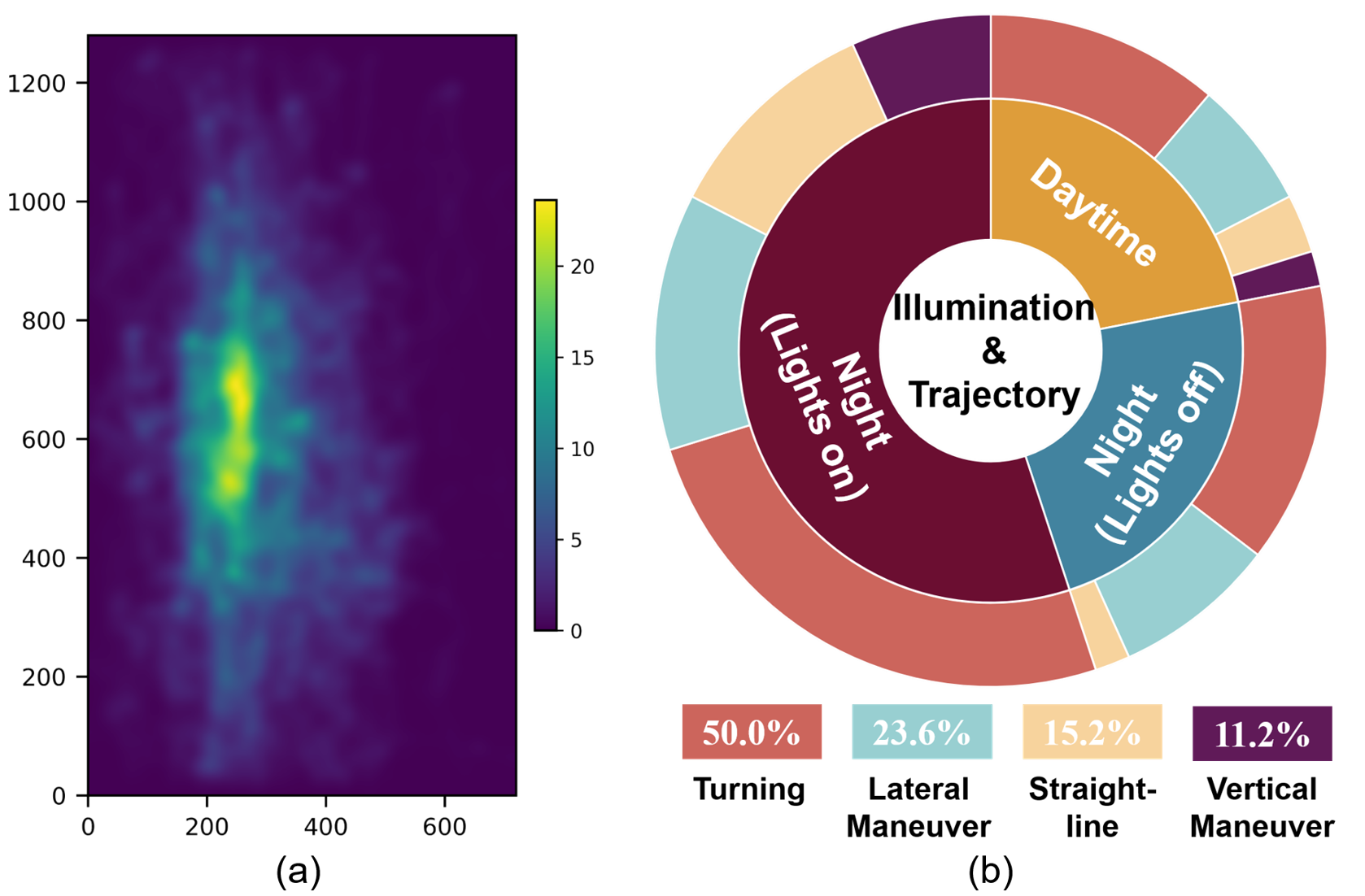}
\caption{Statistics of the proposed dataset. (a) Spatial heatmap of target center distribution. (b) Joint distribution of trajectory patterns and illumination.}
\label{fig:statistics}
\end{figure}

Fig.~\ref{fig:statistics}(a) shows the spatial distribution of target centers, which predominantly appear in the central and upper regions. Small targets in $[16^2, 32^2)$~px$^2$ constitute 57.3\%, and medium targets in $[32^2, 64^2)$~px$^2$ account for 40.1\%. The remaining 2.6\% comprises tiny and large targets. Fig.~\ref{fig:statistics}(b) presents the joint distribution of trajectory patterns and illumination. The dataset emphasizes nighttime sequences to reflect real-world counter-UAV surveillance requirements in A2A operations. Turning maneuvers are the most frequent trajectory pattern across all conditions, followed by lateral and straight-line maneuvers.

\subsection{Splits and Tasks}
\label{sec:splits}

The dataset is partitioned into training, validation, and test subsets containing 125, 18, and 35 sequences, respectively. Two-level hierarchical stratified sampling based on motion geometry and target scale ensures balanced representation, with each subset containing sequences from all major scenario categories. The training set is provided for methods that require supervised learning, while the test set is reserved for final evaluation to prevent overfitting.

\textit{Detection.} Each sequence is divided into non-overlapping slices. The ground-truth box of a slice is the spline evaluated at the end of the slice. Detectors are scored with COCO average precision~\cite{coco} at IoU 0.50 (AP$_{50}$), at IoU 0.75 (AP$_{75}$), and averaged over IoU thresholds from 0.50 to 0.95 (mAP). We keep all boxes and apply no minimum size.

\textit{Tracking.} We follow one-pass evaluation~\cite{otb2015}. A tracker receives event frames at an aggregation rate $f$, each accumulating events over $1/f$, and is initialized with the ground-truth box of the first frame. Its output for a frame is compared with the value of the spline at that frame. We report the area under the success plot (SR), the precision at 20~px (PR)~\cite{otb2015}, the normalized precision at 0.2 (NPR)~\cite{lasot}, and the inference speed in frames per second (FPS).

\section{Benchmark Experiments}
\label{sec:experiments}

\subsection{Setup}
\label{sec:setup}

All baselines use their official implementations and are trained on the training sequences with their default schedules. Experiments run on an Intel Core i5-12400F CPU and an NVIDIA GeForce RTX~3090 GPU. Detectors take 50~ms slices in the input representation required by each method. Trackers are trained and tested at an aggregation rate of 30~Hz unless stated otherwise, so that each frame spans 33.3~ms.

\subsection{Detection Benchmark}
\label{sec:exp-det}

\begin{table}[!t]
\centering
\caption{Detection Results on AE-UAV }
\label{tab:detection}
\setlength{\tabcolsep}{4pt}
\resizebox{\columnwidth}{!}{%
\begin{tabular}{l l l c c c c}
\toprule
\textbf{Method} & \textbf{Source} & \textbf{Rep.} & \textbf{AP$_{50}$}$\uparrow$ & \textbf{AP$_{75}$}$\uparrow$ & \textbf{mAP}$\uparrow$ & \textbf{Runtime}$\downarrow$ \\
\midrule
YOLOX-S~\cite{yolox}    & arXiv21 & Frame  & 65.31 & 24.33 & 30.02 & \textbf{4.9~ms}  \\
RT-DETR-L~\cite{rtdetr} & CVPR24  & Frame  & 62.80 & 23.07 & 29.31 & 18.1~ms \\
GET~\cite{get}          & ICCV23  & Token  & 75.33 & 28.01 & 35.21 & 13.1~ms \\
RVT~\cite{rvt}          & CVPR23  & Hist.  & 75.91 & 28.35 & 35.67 & 8.6~ms  \\
S5-ViT~\cite{s5vit}     & CVPR24  & Hist.  & \underline{76.26} & \underline{28.68} & \underline{35.84} & \underline{6.5~ms} \\
SMamba~\cite{smamba}    & AAAI25  & Voxel  & \textbf{76.72} & \textbf{30.17} & \textbf{36.35} & 15.6~ms \\
\bottomrule
\end{tabular}}
\par\vspace{2pt}{\begin{minipage}{\columnwidth}
\footnotesize
Rep. = event representation: Frame = event frame; Token = group token; Hist. = stacked histogram; Voxel = voxel grid. AP in \%. Bold: best; underline: second best.
\end{minipage}}
\end{table}

Table~\ref{tab:detection} reports the detection results. The four event-specific detectors outperform the two generic frame detectors applied to event frames, suggesting that temporal memory helps separate the target from the moving background. SMamba achieves the highest AP$_{50}$, AP$_{75}$, and mAP, and S5-ViT is the fastest event-specific detector at 6.5~ms per slice. AP$_{75}$ is substantially lower than AP$_{50}$ for every method. Precise localization of small targets is therefore the main weakness.

\begin{table}[!t]
\centering
\caption{Detection mAP (\%) Across Scenarios}
\label{tab:det_scenario}
\setlength{\tabcolsep}{3pt}
\resizebox{\columnwidth}{!}{%
\begin{tabular}{l c c c | c c c}
\toprule
\multirow{2}{*}{\textbf{Method}} & \multicolumn{3}{c|}{\textbf{Illumination}} & \multicolumn{3}{c}{\textbf{Motion geometry}} \\
 & Lights on & Lights off & Daytime & Pursuit & Evasion & Head-on \\
\midrule
YOLOX-S    & 25.09 & 31.37 & 47.98 & 30.62 & 34.54 & 14.55 \\
RT-DETR-L  & 24.74 & 27.69 & 48.72 & 29.47 & 33.62 & 15.94 \\
GET        & 31.56 & 32.23 & 52.64 & 37.05 & 37.64 & 22.19 \\
RVT        & 31.92 & 32.55 & \underline{53.64} & \textbf{38.03} & 37.21 & 23.54 \\
S5-ViT     & \underline{32.06} & \underline{32.76} & \textbf{53.88} & 37.47 & \underline{38.15} & \underline{23.81} \\
SMamba     & \textbf{32.38} & \textbf{34.51} & 53.39 & \underline{37.63} & \textbf{39.17} & \textbf{23.86} \\
\bottomrule
\end{tabular}}
\par\vspace{2pt}{\begin{minipage}{\linewidth}
\footnotesize
Lights on and Lights off refer to nighttime sequences with the target lights on or off. Bold: best; underline: second best.
\end{minipage}}
\end{table}

Table~\ref{tab:det_scenario} breaks down mAP by illumination and motion geometry. Daytime sequences yield the highest mAP for all methods. Nighttime sequences are considerably harder. Across all detectors, the lights-off condition is slightly easier than lights-on. A likely cause is that flickering artificial light sources resemble the target navigation lights and act as distractors in lights-on sequences. Generic frame detectors suffer the most from this confusion, and both fall to an mAP of about 25\% under lights-on. Evasion yields the highest mAP among the three geometries for most event-specific detectors. Head-on encounters present the greatest challenge. All methods achieve an mAP below 24\% in this geometry, roughly half their evasion scores. Limited training data for this geometry further compounds the difficulty, as only 19 sequences involve head-on flight. The margin of the event-specific detectors over the generic frame detectors is largest in this geometry. Nighttime and head-on encounters are thus the two A2A conditions that current detectors handle worst.

\subsection{Tracking Benchmark}
\label{sec:exp-track}

\begin{table}[!t]
\centering
\caption{Single-Object Tracking Results on AE-UAV at 30~Hz}
\label{tab:tracking}
\small
\resizebox{\columnwidth}{!}{%
\begin{tabular}{l l c c c c}
\toprule
\textbf{Method} & \textbf{Source} & \textbf{SR}$\uparrow$ & \textbf{PR}$\uparrow$ & \textbf{NPR}$\uparrow$ & \textbf{FPS}$\uparrow$ \\
\midrule
OSTrack~\cite{ostrack}       & ECCV22  & 32.95 & 59.20 & 38.67 & \textbf{148} \\
ROMTrack~\cite{romtrack}     & ICCV23  & 55.82 & 86.84 & 75.32 & \underline{126} \\
AQATrack~\cite{aqatrack}     & CVPR24  & \underline{56.07} & \textbf{88.52} & \underline{80.23} & 112 \\
ODTrack~\cite{odtrack}       & AAAI24  & 35.24 & 64.25 & 42.68 & 30 \\
MambaEVT~\cite{mambaevt}     & TCSVT25 & 40.56 & 65.38 & 40.86 & 20 \\
FocusTrack~\cite{focustrack} & TGRS25  & \textbf{60.05} & \underline{88.37} & \textbf{81.63} & 122 \\
\bottomrule
\end{tabular}}
\par\vspace{2pt}{\begin{minipage}{\linewidth}
\footnotesize
SR, PR, and NPR are in \%. Bold: best; underline: second best.
\end{minipage}}
\end{table}

Table~\ref{tab:tracking} reports the tracking results at 30~Hz. FocusTrack, designed for counter-UAV tracking, achieves the highest SR of 60.05\% and the highest NPR of 81.63\%. AQATrack ranks second in SR at 56.07\% and leads in PR at 88.52\%, with ROMTrack close behind at 55.82\% SR. The SR of trackers built on similar transformer backbones ranges from 32.95\% to 60.05\%, so the benchmark separates methods clearly. OSTrack is the fastest at 148~FPS but is also the least accurate, whereas ROMTrack and FocusTrack combine an SR above 55\% with more than 120~FPS.

\begin{table}[!t]
\centering
\caption{Tracking SR (\%) Across Scenarios}
\label{tab:track_scenario}
\setlength{\tabcolsep}{3pt}
\resizebox{\columnwidth}{!}{%
\begin{tabular}{l c c c | c c c}
\toprule
\multirow{2}{*}{\textbf{Method}} & \multicolumn{3}{c|}{\textbf{Illumination}} & \multicolumn{3}{c}{\textbf{Motion geometry}} \\
 & Lights on & Lights off & Daytime & Pursuit & Evasion & Head-on \\
\midrule
OSTrack    & 33.56 & 31.14 & 33.32 & 24.34 & 45.24 & 23.64 \\
ROMTrack   & 56.22 & \underline{55.98} & 54.03 & 55.79 & 59.72 & \underline{44.25} \\
AQATrack   & \underline{56.78} & 55.26 & \underline{54.64} & \underline{55.96} & \underline{60.11} & 44.21 \\
ODTrack    & 35.88 & 33.27 & 35.39 & 26.72 & 47.54 & 25.65 \\
MambaEVT   & 42.41 & 37.38 & 37.97 & 37.44 & 48.37 & 26.77 \\
FocusTrack & \textbf{59.34} & \textbf{61.94} & \textbf{59.86} & \textbf{60.97} & \textbf{62.13} & \textbf{47.45} \\
\bottomrule
\end{tabular}}
\par\vspace{2pt}{\begin{minipage}{\linewidth}
\footnotesize
Lights on and Lights off refer to nighttime sequences with the target lights on or off. Bold: best; underline: second best.
\end{minipage}}
\end{table}

Table~\ref{tab:track_scenario} breaks the tracking results down by scenario. 
Unlike in detection, illumination has a relatively mild effect on the top trackers. Among the illumination conditions, FocusTrack achieves its highest SR of 61.94\% under the lights-off condition. A tracker is initialized on the target and searches a local region around its last position, whereas a detector must find the target in the whole slice, where nighttime noise produces false candidates. Evasion sequences produce the highest SR for all trackers, consistent with the detection. Head-on encounters remain the hardest geometry. Even FocusTrack reaches only 47.45\% SR in head-on encounters, roughly 15 percentage points below its evasion score. 

\subsection{Cross-Rate Generalization of Trackers}
\label{sec:exp-rate}

\begin{figure}[!t]
\vspace*{5pt}
\centering
\includegraphics[width=0.95\columnwidth]{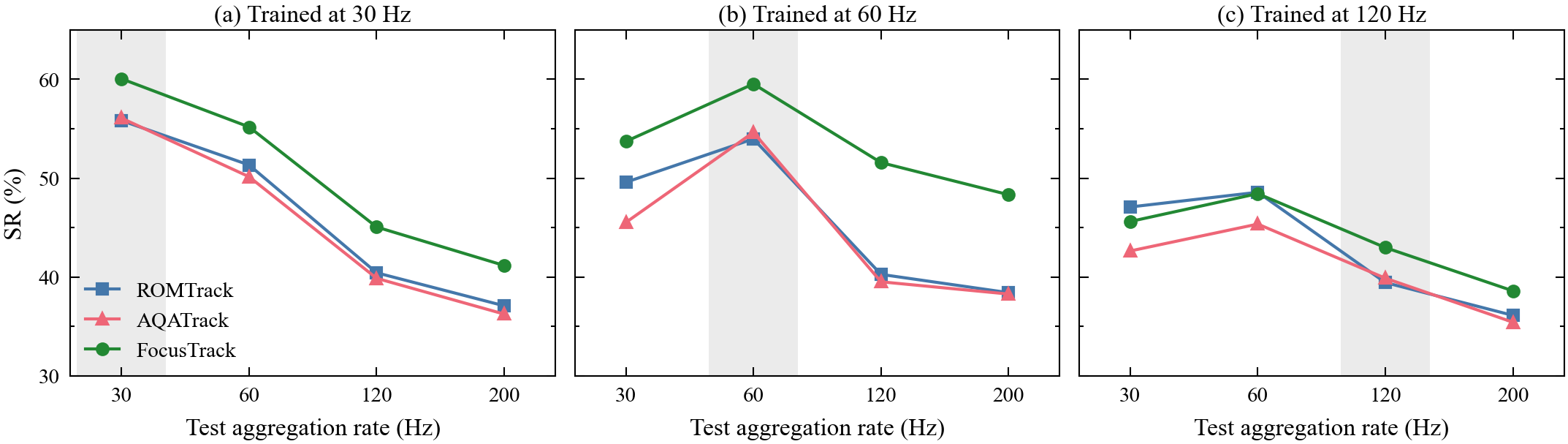}
\caption{SR of three learned trackers across test aggregation rates. Each panel corresponds to one training rate, and the shaded column marks the test rate that equals the training rate.}
\label{fig:rate}
\end{figure}

Unlike frame cameras, which deliver a fixed frame rate, event cameras allow flexible temporal aggregation, and a tracker deployed on a real platform may need to operate at varying rates depending on latency budgets or computational constraints.
Fig.~\ref{fig:rate} examines how the three best trackers generalize across aggregation rates. In each panel, all trackers are trained at a single rate and evaluated at test rates of 30, 60, 120, and 200~Hz; the shaded column marks the training rate.

All three trackers degrade mainly when the test rate exceeds the training rate.
When trained at 30~Hz, FocusTrack drops from 60.05\% to 41.16\% at 200~Hz, losing roughly one third of its accuracy. ROMTrack and AQATrack exhibit similar proportional losses, falling from approximately 56\% to 37\%. Higher aggregation rates shorten the temporal window of each frame, reducing the number of events available per frame and weakening the appearance cues that learned templates rely on. This effect is most severe at 200~Hz, where each frame spans only 5~ms. 

Training at a higher rate does not solve the problem. Trained at 60~Hz, the trackers peak at 60~Hz but are less accurate at 30~Hz than the models trained at 30~Hz.  Trained at 120~Hz, all three trackers perform better at 30 and 60~Hz than at their own rate. No single training rate dominates across all test rates. Bridging this rate gap remains an open challenge for event-based tracking, and AE-UAV provides the labels needed to measure it.

\section{Limitations}
\label{sec:limitations}

AE-UAV has three limitations. First, it contains a single target type observed by one platform with a fixed lens, so appearance diversity comes only from viewpoint and illumination rather than airframe variation. Second, the dataset provides event data alone, precluding evaluation of methods that fuse inertial measurements for ego-motion compensation. Third, per-event labels are derived from axis-aligned bounding boxes. Background events that fall inside the box are labeled as target events. Methods that rely on per-event labels may learn from noisy supervision near the box boundary. Future extensions should include additional target UAV models to test whether learned detectors and trackers generalize across platforms.

\section{Conclusion}
\label{sec:conclusion}
We presented AE-UAV, an A2A event camera dataset for UAV detection and tracking. It comprises 178 sequences spanning daytime and nighttime conditions under three motion geometries. A cubic B-spline annotation framework provides continuous-time ground truth at arbitrary timestamps. Benchmark evaluation shows that head-on encounters are the hardest geometry for both detection and tracking, whereas nighttime mainly degrades detection. When trained at 30~Hz, the three strongest trackers lose roughly one third of their SR at 200~Hz, and no training rate is best at all test rates. These findings identify rate robustness as an open challenge for event-based tracking. We hope AE-UAV will support research on event-driven A2A perception.


\end{document}